%% file: main.tex
\documentclass[runningheads]{llncs}

\usepackage{eccv}

\usepackage{eccvabbrv}

\usepackage{graphicx}
\usepackage{booktabs}

\usepackage[accsupp]{axessibility}  

\usepackage[pagebackref,breaklinks,colorlinks,citecolor=eccvblue]{hyperref}

\usepackage{orcidlink}
\usepackage{bm}
\usepackage{amsmath}
\usepackage{multirow}

\def\blue#1{\textcolor{blue}{#1}}
\newcommand{\ourproblem}{\texttt{U}$^2$\texttt{CDR}}
\newcommand{\IDSE}{\texttt{IDSE}}
\newcommand{\CDSM}{\texttt{CDSM}}
\newcommand{\IPM}{\texttt{IPM}}
\newcommand{\SEL}{\texttt{SEL}}
\newcommand{\SPDA}{\texttt{SPDA}}
\newcommand{\SNNM}{\texttt{SN$^2$M}}
\newcommand{\method}{\texttt{UEM}}

\begin{document}

\title{Semantic Feature Learning for Universal Unsupervised Cross-Domain Retrieval} 


\author{Lixu Wang\inst{1} \and
Xinyu Du\inst{2} \and
Qi Zhu\inst{1}}

\authorrunning{L. Wang et al. Preprint}

\institute{Northwestern University, Evanston IL, USA \and
General Motors Global R\&D, MI, USA
\\ \email{\{lixuwang2025@u., qzhu@\}northwestern.edu}\\
\email{xinyu.du@gm.com}}

\maketitle

\begin{abstract}
  Cross-domain retrieval (CDR), as a crucial tool for numerous technologies, is finding increasingly broad applications. However, existing efforts face several major issues, with the most critical being the need for accurate supervision, which often demands costly resources and efforts. Cutting-edge studies focus on achieving unsupervised CDR but typically assume that the category spaces across domains are identical, an assumption that is often unrealistic in real-world scenarios. This is because only through dedicated and comprehensive analysis can the category spaces of different domains be confirmed as identical, which contradicts the premise of unsupervised scenarios. Therefore, in this work, we introduce the problem of \underline{U}niversal \underline{U}nsupervised \underline{C}ross-\underline{D}omain \underline{R}etrieval (\ourproblem{}) for the first time and design a two-stage semantic feature learning framework to address it. In the first stage, a cross-domain unified prototypical structure is established under the guidance of an instance-prototype-mixed contrastive loss and a semantic-enhanced loss, to counteract category space differences. In the second stage, through a modified adversarial training mechanism, we ensure minimal changes for the established prototypical structure during domain alignment, enabling more accurate nearest-neighbor searching. Extensive experiments across multiple datasets and scenarios, including closet, partial, and open-set CDR, demonstrate that our approach significantly outperforms existing state-of-the-art CDR works and some potentially effective studies from other topics in solving \ourproblem{} challenges.
\end{abstract}

\input{Sections/Introduction}
\input{Sections/RelatedWork}

\input{Sections/Methodology}

\input{Sections/Experiments}
\input{Sections/Conclusion}

%
%
\bibliographystyle{splncs04}
\bibliography{main}
\end{document}

%% file: Sections/Introduction.tex
\section{Introduction}
In real-world applications, cross-domain retrieval finds extensive utility, spanning diverse domains such as image search~\cite{imagesearch}, product recommendations~\cite{recommendation}, and artistic creation~\cite{styletransfer, styletransfer1}. However, the efficacy of current cross-domain retrieval methods relies heavily on accurate and sufficient supervision~\cite{cdir1, cdir2}, whether category labels or cross-domain pairing information. The acquisition of such supervision demands costly efforts and resources. Hence, there is an urgent need to develop unsupervised cross-domain retrieval (UCDR) techniques.

\begin{figure*}[htbp]
\centering
\includegraphics[width=1.\textwidth]{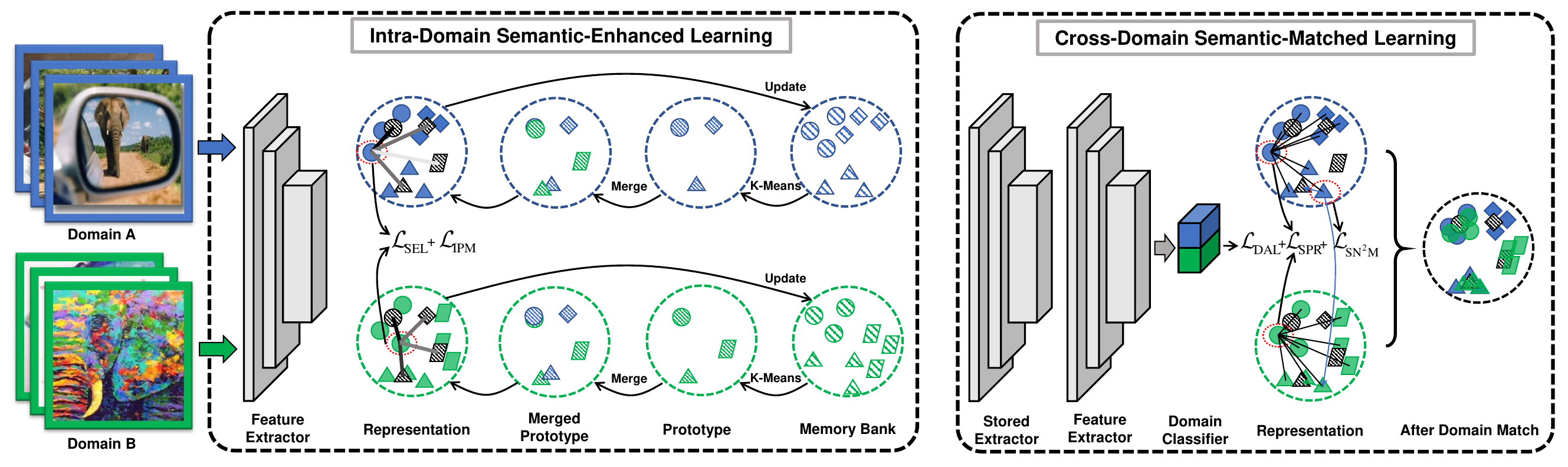}
\vspace{-20pt}
\caption{Overview of our proposed \method{} semantic feature learning framework for \ourproblem{}. In the first stage of Intra-Domain Semantic-Enhanced Learning, \method{} establishes a unified prototypical structure across domains, which is driven and enhanced by \IPM{} and \SEL{}, respectively. Then, in the second stage of Cross-Domain Semantic-Matched Learning, \SPDA{} is used to align domains while preserving the built prototypical structure, and \SNNM{} can achieve more accurate cross-domain categorical pairing.}
\label{fig_overview}
\vspace{-15pt}
\end{figure*}

For the regular UCDR problem~\cite{ucdr1, ucdr2}, there are two data domains with semantic similarity but distinct characteristics: one is the query domain, while the other is the retrieval domain. Despite the absence of category labels, regular UCDR typically assumes that the label spaces of both domains are identical. However, in real-world applications~\cite{cdir1}, \textbf{\textit{the categorical composition of an unlabeled data domain is usually uncertain, which is hard to acquire without detailed analysis and dedicated expertise}}. In this work, we focus on extending UCDR to more universal scenarios, which allow for the possibility of disparate category spaces across domains. The objective of \emph{Universal UCDR} (\ourproblem{}) is to retrieve samples from the retrieval domain that share the same category label with a query sample from the query domain. Naturally, in the case of private categories exclusive to the query domain, the retrieval result should be null.

Two challenges must be tackled when solving traditional UCDR: 1) effectively distinguishing data samples in each domain without category labels, and 2) achieving alignment across domains for samples of the same category without cross-domain pairing information. For the first challenge, existing approaches typically employ self-supervised learning~\cite{infoNCE, moco} (SSL) independently within each domain. As for the second, they often rely on diverse nearest-neighbor searching algorithms. However, for \ourproblem{}, applying these existing works introduces new issues. First, the prevailing SSL methods, particularly contrastive learning~\cite{infoNCE, ProtoNCE, moco}, is highly influenced by the category label space~\cite{partial_CL}, which means different label spaces lead to distinct semantic structures that hinder the subsequent cross-domain category alignment. Moreover, existing nearest-neighbor searching algorithms~\cite{PCS, CDS2021, ucdr1} overlook the presence of domain gaps. We found that only by first addressing the domain gap can the nearest neighbor searching based on cross-domain semantic similarity become reliable and accurate.

Thus, to effectively address the challenges of \ourproblem{}, we propose a two-stage Unified, Enhanced, and Matched (\method{}) semantic feature learning framework. In the first stage, we establish a cross-domain unified prototypical structure with an instance-prototype-mixed (\IPM{}) contrastive loss, accompanied by a semantic-enhanced loss (\SEL{}) to guide the model towards this structure during training. Then, before conducting cross-domain category alignment, we incorporate Semantic-Preserving Domain Alignment (\SPDA{}) to diminish the domain gap while ensuring minimal changes for the established prototypical structure. As the domain gap diminishes, we propose a more effective nearest-neighbor searching algorithm -- Switchable Nearest Neighboring Match (\SNNM{}), to select more reliable cross-domain neighbors based on the relationship between instances and prototypes. Extensive experiments and ablation studies on popular benchmark datasets demonstrate that our work can substantially outperform state-of-the-art works from Cross-Domain Representation Learning, Unsupervised Domain Generalization, and UCDR in cases where the label spaces cross domains are the same or not. In summary, our major contributions are:
\begin{itemize}
    \item We are the first to identify and solve an important problem when employing unsupervised cross-domain retrieval (UCDR) in the real world -- Universal UCDR (\ourproblem{}) where the category spaces of different domains are distinct.
    \item We propose a two-stage Unified, Enhanced, and Matched (\method{}) semantic feature learning framework to solve \ourproblem{}. In the first stage, \method{} tries to establish a unified prototypical structure across domains, which can ensure consistent semantic learning even with category space differences. Then, \method{} can achieve effective domain alignment and cross-domain pairing while enabling minimal changes for the prototypical structure.
    \item We conduct extensive experiments on multiple benchmark datasets with settings including Closet, Partial, and Open-set UCDR. The results demonstrate that \method{} can substantially outperform state-of-the-art works of UCDR and other potential solutions in all settings.
\end{itemize}

%% file: Sections/RelatedWork.tex
\section{Related Work}
\noindent \textbf{Cross-Domain Image Retrieval.}
Image retrieval aims to retrieve similar images for a query image~\cite{smeulders2000content}, and its cross-domain extension considers scenarios where the distributions of the query images and the retrieval database are distinct~\cite{cdr}. Cross-domain retrieval (CDR) is conducted at the category level, i.e., for query samples belonging to a particular category, it is ideal that the retrieved samples belong to the same category. This objective is not very difficult to achieve if there are categorical labels~\cite{cdr, cdr_survey}. However, in real-world applications, such categorical labeling information is hard to acquire, thus more recent works~\cite{ucdr1, ucdr2, coda} spare no effort to achieve unsupervised CDR (UCDR). CDS~\cite{CDS2021} makes the first step and proposes a contrastive learning-based cross-domain pre-training to align different domains. PCS~\cite{PCS} discovers some issues in CDS and incorporates prototype contrastive learning~\cite{ProtoNCE} into the cross-domain pre-training. Recent studies also search for various ways like clustering~\cite{ucdr1}, pseudo-labeling~\cite{ge2024pseudo}, and classifier mixup~\cite{coda}, to achieve more advanced cross-domain match after or during the unsupervised pre-training. Besides, there are also some works~\cite{ucdr2} leveraging data augmentation to achieve domain-generalizable UCDR. However, to our best knowledge, all these UCDR works are built on the assumption that both the query and retrieval domains share the same category space, which is impractical in the real world. By contrast, we focus on solving the problem of universal UCDR in this work, which allows image retrieval to work effectively when the category space is distinct.

\smallskip \noindent \textbf{Universal Cross-Domain Learning.}
Cross-domain learning usually consists of domain adaptation (DA)~\cite{xu2021weak} and domain generalization (DG)~\cite{dg_survey, mohapatra2024phase}. In DA, two domains of data are given, but one is labeled (source) and another is unlabeled (target), and the objective is to train a model for the target domain. As for DG, the target domain is unavailable, but the objective is still the same. Regular DA and DG also only consider the scenario where the label space of the target domain is the same as the source label space, which is termed Closet DA/DG. As the research about real-world applications deepens, more studies realize that the target label space may be a subset of the source one (Partial DA/DG)~\cite{pda1, pda2} or contain some private labels that other domains don't have (Openset DA/DG)~\cite{openda1, openda2}. To deal with such universal setups, UniDA~\cite{unida} proposes the first framework to unify entropy and domain similarity for quantifying sample transferability across domains and allocate different learning weights to samples. CMU~\cite{cmu} extends the transferability quantification into three metrics -- entropy, consistency, and confidence. More recent works search ways like clustering~\cite{newuda3, newuda4} and nearest neighbor matching~\cite{newuda1, newuda2, wang2021providing} to achieve universal DA/DG. In addition, some studies appear to achieve unsupervised DG where the source domain is also unlabeled~\cite{darling, DN2A}, which is similar to the setup of UCDR. The representative method is to achieve consistency across different data augmentations. Besides, considering cost-cutting and privacy protection~\cite{wang2022non, guo2024domain}, more recent works have found ways to achieve source-free~\cite{wang2021providing} and continual DA/DG~\cite{liu2022deja}. However, these studies consider the classification task with at least some supervision information available, usually from the source domain. They cannot effectively work in image retrieval, especially in completely unsupervised cases.

%% file: Sections/Methodology.tex
\section{Methodology}
\subsection{Problem Formulation}
In the problem of \ourproblem{}, we assume there are two domains characterized by $N^\mathrm{A}$ and $N^\mathrm{B}$ unlabeled instances, which are denoted as $\mathcal{D}^\mathrm{A}=\{\bm{x}_i^\mathrm{A}\}_{i=1}^{N^\mathrm{A}}$ and $\mathcal{D}^\mathrm{B}=\{\bm{x}_i^\mathrm{B}\}_{i=1}^{N^\mathrm{B}}$, respectively. Although these two domains are provided as unlabeled data without category labels, we assume their label spaces $\mathcal{Y}^\mathrm{A}, \mathcal{Y}^\mathrm{B}$ consist of $C^\mathrm{A}$ and $C^\mathrm{B}$ different categories, and there is a relationship that $C^\mathrm{A} \neq C^\mathrm{B}, \mathcal{Y}^\mathrm{A} \cap \mathcal{Y}^\mathrm{B} \neq \mathcal{Y}^\mathrm{A} \cup \mathcal{Y}^\mathrm{B}$. Without losing generality, if we regard domain A as the query domain, while domain B is the retrieval domain, the objective of \ourproblem{} is to retrieve correct data from domain B that belongs to the same categories as the query data provided by domain A. To achieve this objective, it is required to train a valid feature extractor $f_\theta: \mathcal{X} \rightarrow \mathcal{R}$ that can map both these two domains from the input space $\mathcal{X}$ to a feature space $\mathcal{R}$. Then the retrieval process $R(f_\theta, \bm{x}_i^\mathrm{A})$ is shaped like for a particular query instance $\bm{x}_i^\mathrm{A}$ with the label $y_i^\mathrm{A}$ from domain A, the representation distance between all instances in domain B and $\bm{x}_i^\mathrm{A}$ needs to be calculated to form a set, i.e., $\mathcal{S} = \{\mathrm{d}(f(\bm{x}_j^\mathrm{B}), f(\bm{x}_i^\mathrm{A}))\}_{j=1}^{N^\mathrm{B}}$ where $\mathrm{d}(\cdot)$ is a particular distance metric (e.g., Euclidean Distance), and we have

\begin{equation}
R(f_\theta, \bm{x}_i^\mathrm{A}) = \left\{ 
\begin{array}{ll}
\mathrm{null},\, \text{if}\, y_i^\mathrm{A} \in \mathcal{Y}^\mathrm{A} \setminus \mathcal{Y}^\mathrm{B}\\
\mathrm{sort}_\uparrow(\mathcal{S})[1:k],\, \text{otherwise},
\end{array}
\right.
\end{equation}
where $\mathrm{sort}_\uparrow(\cdot)$ means to sort the input set in ascending order, and $[1:k]$ denotes the first $k$ elements of the sorted set.

\smallskip \noindent \textbf{Method Overview.}
To solve \ourproblem{}, we propose a Unified, Enhanced, and Matched (\method{}) semantic feature learning framework that consists of two stages -- Intra-Domain Semantic-Enhanced (\IDSE{}, Section~\ref{sec_IDSE}) Learning and Cross-Domain Semantic-Matched (\CDSM{}, Section~\ref{sec_CDSM}) Learning, which is shown in Figure~\ref{fig_overview}. \IDSE{} can help the feature extractor $f_\theta$ to extract categorical semantics and ensure a consistent semantic structure across domains at the same time, which is achieved by instance-prototype-mixed (\IPM{}, Section~\ref{sec_ipm}) contrastive learning and a novel Semantic-Enhanced Loss (\SEL{}, Section~\ref{sec_sel}). After \IDSE{}, \CDSM{} conducts Semantic-Preserving Domain Alignment (\SPDA{}, Section~\ref{sec_spda}) to minimize the domain gap while preserving the semantic structure learned by \IDSE{}. With the minimization of the domain gap, more accurate nearest-neighbor searching can be achieved by our Switchable Nearest Neighboring Match (\SNNM{}, Section~\ref{sec_SNNM}).

\subsection{Intra-Domain Semantic-Enhanced Learning}
\label{sec_IDSE}
To achieve effective cross-domain retrieval, feature extractor $f_\theta$ needs to learn consistent cross-domain features to differentiate data categories. Instance Discrimination~\cite{infoNCE} is usually employed to achieve discriminative feature learning, but solely applying it in \ourproblem{} has four fundamental issues that hinder the possibility of accurate cross-domain categorical matching later:

\noindent 1) Instance discrimination tends to extract semantics that separate domains rather than categories, e.g.,
\begin{equation}
\small
    \mathrm{d}(f(\bm{x}_i^\mathrm{A}), f(\bm{x}_j^\mathrm{A})) < \mathrm{d}(f(\bm{x}_i^\mathrm{A}), f(\bm{x}_j^\mathrm{B})), \, y_i^\mathrm{A} = y_j^\mathrm{B} \neq y_j^\mathrm{A}.
    \label{eq_issue_1}
\end{equation}
2) Instance discrimination cannot encode the categorical semantics in the input space into the feature space, e.g.,
\begin{equation}
\small
    \frac{\mathrm{d}(\bm{x}_i^\mathrm{A}, \bm{x}_{j_1}^\mathrm{A})}{\mathrm{d}(\bm{x}_i^\mathrm{A}, \bm{x}_{j_2}^\mathrm{A})} < \frac{\mathrm{d}(f(\bm{x}_i^\mathrm{A}), f(\bm{x}_{j_1}^\mathrm{A}))}{\mathrm{d}(f(\bm{x}_i^\mathrm{A}), f(\bm{x}_{j_2}^\mathrm{A}))}, \,y_i^\mathrm{A} = y_{j_1}^\mathrm{A} \neq y_{j_2}^\mathrm{A}.
    \label{eq_issue_2}
\end{equation}
3) The randomness introduced by stochastic data augmentations results in evident changes in learned categorical semantic structures during training, i.e.,
\begin{equation}
\small
    \mathrm{d}(G(\mathcal{P}_t^\mathrm{A}), G(\mathcal{P}_{t+1}^\mathrm{A})) \gg \min\limits_{\mathcal{P}_i^\mathrm{A}, \mathcal{P}_j^\mathrm{A} \sim \mathcal{H}} \mathrm{d}(G(\mathcal{P}_i^\mathrm{A}), G(\mathcal{P}_j^\mathrm{A})),
    \label{eq_issue_3}
\end{equation}
where $G(\cdot)$ corresponds to a graph constructed by the input vectors, and $\mathcal{P}$ denotes the set of categorical prototypes for a domain. The subscript $t$ denotes different training iterations, while $\mathcal{H}$ represents the hypothesis space of possible categorical prototype sets for a particular domain. $\mathrm{d}(\cdot)$ here is a measurement for graph difference, e.g., graph edit distance~\cite{ged}.

\noindent 4) Distinct label spaces make instance discrimination learn different categorical semantic structures, i.e.,
\begin{equation}
\small
    \mathrm{d}(G(\mathcal{P}^\mathrm{A}), G(\mathcal{P}^\mathrm{B})) \gg \min\limits_{\mathcal{P}_i^\mathrm{A} \sim \mathcal{H}^\mathrm{A}, \mathcal{P}_j^\mathrm{B} \sim \mathcal{H}^\mathrm{B}} \mathrm{d}(G(\mathcal{P}_i^\mathrm{A}), G(\mathcal{P}_j^\mathrm{B})).
    \label{eq_issue4}
\end{equation}

\subsubsection{Instance-Prototype-Mixed Contrastive Learning.}
\label{sec_ipm}
To fix the above issues, we adopt an augmentation-free contrastive learning algorithm -- MoCo~\cite{moco}, to handle the third issue reflected by Eq.~(\ref{eq_issue_3}). Moreover, the MoCo-based instance discrimination is conducted separately for each domain, which encourages $f_\theta$ to focus less on learning domain semantics (for the first issue, Eq.~(\ref{eq_issue_1})). Besides, we accompany MoCo with a prototypical contrastive loss, which is also augmentation-free, to enhance the mapping of categorical semantics from the input space to the feature space (for addressing the second issue, Eq.~(\ref{eq_issue_2})). With a well-crafted prototype update mechanism, this prototypical contrastive loss can also help build a consistent semantic structure across domains (for the last issue, Eq.~(\ref{eq_issue4})). Then let us introduce \IPM{} in detail. First of all, two memory banks $\mathcal{M}^\mathrm{A}$ and $\mathcal{M}^\mathrm{B}$ are maintained for samples from domains A and B:
\begin{align}
\mathcal{M}^\mathrm{A} = \left[\bm{m}_1^\mathrm{A}, ..., \bm{m}_{N^\mathrm{A}}^\mathrm{A}\right], \mathcal{M}^\mathrm{B}= \left[\bm{m}_1^\mathrm{B}, ..., \mathbf{m}_{N^\mathrm{B}}^\mathrm{B}\right],   
\end{align}
where $\bm{m}_i$ is the stored historical feature of data instance $\bm{x}_i$. $\bm{m}_i$ is initialized by the feature of $\bm{x}_i$ extracted by the initial $f_\theta$ and updated in momentum,
\begin{align}
\bm{m}_i \leftarrow \beta \bm{m}_i + (1-\beta)f_\theta(\bm{x}_i),
\end{align}
where $\beta$ controls the momentum speed, and we set it as a popular value $0.99$. With these two memory banks, MoCo builds the positive pairs as the pair of each instance and its historical feature, while the negative ones are pairs of each instance and the historical features of all other instances. The instance discrimination is conducted as
\begin{align}
\mathcal{L}_\mathrm{INCE} = \sum_{i=1}^{B} -\log \frac{\exp{(f_\theta(\bm{x}_i) \cdot \bm{m}_{i}/\tau)}}{\sum_{j=1}^{B}\exp{(f_\theta(\bm{x}_i) \cdot \bm{m}_{j}/\tau)}},
\end{align}
where $B$ is the batch size and $\tau$ is a temperature factor that is set as $0.07$.

As for the design of our prototypical contrastive loss, K-Means is applied on $\mathcal{M}^\mathrm{A}$ and $\mathcal{M}^\mathrm{B}$ to construct prototypes as cluster centers $\mathcal{P} = \{\bm{p}_c\}_{c=1}^{\widehat{C}}$. In our problem, the cluster number $C$ is unknown, thus we apply the Elbow approach~\cite{elbow} to estimate it as $\widehat{C}$. Then, for each instance $\bm{x}_i$, if it belongs to the $c_i$-th cluster, the prototypical contrastive loss $\mathcal{L}_\mathrm{PNCE}$ shapes like,
\begin{align}
\mathcal{L}_\mathrm{PNCE} = \sum_{i=1}^{B} -\log \frac{\exp{(f_\theta(\bm{x}_i) \cdot \bm{p}_{c_i}/\tau)}}{\sum_{c=1}^{\widehat{C}}\exp{(f_\theta(\bm{x}_i) \cdot \bm{p}_{c}/\tau)}}.
\label{eq_pnce}
\end{align}

Until here, the first three issues may be fixed by mixing $\mathcal{L}_\mathrm{INCE}$ and $\mathcal{L}_\mathrm{PNCE}$, but the last issue, Eq.~(\ref{eq_issue4}), is the bottleneck of \ourproblem{}. Next, let us introduce how we build a consistent prototypical structure for $\mathcal{L}_\mathrm{PNCE}$ to address the last issue. Specifically, after conducting K-Means and Elbow estimation on $\mathcal{M}^\mathrm{A}$ and $\mathcal{M}^\mathrm{B}$ respectively, we can obtain the prototype sets $\mathcal{P}^\mathrm{A}, \mathcal{P}^\mathrm{B}$ of domain A and B. Then taking domain A as an example to illustrate the prototypical structure building process, the prototype set $\mathcal{P}^\mathrm{B}$ of domain B will be translated to domain A as,
\begin{align}
    \mathcal{P}^{\mathrm{B}\rightarrow\mathrm{A}} = \{\bm{p}_c^{\mathrm{B}\rightarrow\mathrm{A}} = \overrightarrow{\bm{p}_c^\mathrm{B}} + \overrightarrow{\overline{\mathcal{M}^\mathrm{B}}\,\overline{\mathcal{M}^\mathrm{A}}}\}_{c=1}^{\widehat{C}^\mathrm{B}},
\end{align}
where $\overline{\mathcal{M}}$ denotes the average feature vector of all vectors in $\mathcal{M}$. Next, each prototype $\bm{p}_c^{\mathrm{B}\rightarrow\mathrm{A}}$ in $\mathcal{P}^{\mathrm{B}\rightarrow\mathrm{A}}$ searches its closet prototype $\bm{p}_{c^\prime}^\mathrm{A}$ in $\mathcal{P}^\mathrm{A}$ for the opportunity of merging, which needs to satisfy the condition,
\begin{align}
    d(\bm{p}_c^{\mathrm{B}\rightarrow\mathrm{A}}, \bm{p}_{c^\prime}^\mathrm{A}) < \min\left[\min\limits_{\bm{p}_i, \bm{p}_j \in \mathcal{P}^\mathrm{A}} d(\bm{p}_i, \bm{p}_j), \min\limits_{\bm{p}_i, \bm{p}_j \in \mathcal{P}^\mathrm{B}} d(\bm{p}_i, \bm{p}_j)\right],
    \label{eq_merge_con}
\end{align}
where $d(\cdot, \cdot)$ computes the Euclidean distance between two given vectors. The merging operation is simple vector averaging, and if we use the symbol $\oplus$ to identify prototypes that satisfy this merging condition, the final prototypical structure for domain A is $\mathcal{P}^{\mathrm{A}^\prime} = \left(\mathcal{P}^\mathrm{A} \setminus \mathcal{P}^{\mathrm{A}, \oplus}\right) \cup \left(\mathcal{P}^{\mathrm{B}\rightarrow\mathrm{A}} \setminus \mathcal{P}^{\mathrm{B}\rightarrow\mathrm{A}, \oplus}\right) \cup \left(\mathcal{P}^{\mathrm{A}, \oplus} \oplus \mathcal{P}^{\mathrm{B}\rightarrow\mathrm{A}, \oplus}\right)$ where $\left(\mathcal{P}^{\mathrm{A}, \oplus} \oplus \mathcal{P}^{\mathrm{B}\rightarrow\mathrm{A}, \oplus}\right) = \left\{\left(\bm{p}_c^{\mathrm{B}\rightarrow\mathrm{A}, \oplus} +  \bm{p}_{c^\prime}^{\mathrm{A}, \oplus}\right)/2\right\}_{c=1}^{C^\oplus}$. Then the computation of $\mathcal{L}_\mathrm{PNCE}^{\mathrm{A}^\prime}$ -- Eq.~(\ref{eq_pnce}) for domain A is conducted on the newly-built $\mathcal{P}^{\mathrm{A}^\prime}$, and all these operations are same for domain B.

However, as establishing the semantic cluster-like structure requires time, it is unreasonable to conduct prototype contrastive learning from the beginning of training. Therefore, we conduct instance discrimination at the beginning and progressively incorporate $\mathcal{L}_\mathrm{PNCE}$. In this case, not only the constructed cluster centers are more reliable, but the Elbow approach also provides more accurate cluster number estimations. Specifically, we use a coefficient $\alpha$ that is scheduled by a Sigmoid function to control the incorporation weight of $\mathcal{L}_\mathrm{PNCE}$, i.e., 
\begin{align}
    \mathcal{L}_\mathrm{IPM} = \mathcal{L}_\mathrm{INCE} + \alpha \mathcal{L}_\mathrm{PNCE}, \,\text{where}\, \alpha = \frac{1}{1 + \exp{(0.5E - e)}},
\end{align}
$E$ and $e$ here are the overall training epochs and the current epoch for \IDSE{}.

\subsubsection{Semantic-Enhanced Loss.}
\label{sec_sel}
For the \IPM{} contrastive learning, it is arbitrary to allocate a data instance $\bm{x}_i$ to a single cluster when the preferred semantic prototypical structure cannot be learned in advance. As a result, to speed up the structure-building process, we propose a novel \emph{Semantic-Enhanced Loss} (\SEL{}) to better align data instances with the prototypes. Specifically, instead of assigning data instances with a single cluster, \SEL{} considers potential semantic relationships between instances with all clusters, which are measured by the Softmax probability. Moreover, as both $\mathcal{P}^\mathrm{A}$ and $\mathcal{P}^\mathrm{B}$ are obtained by the Euclidean distance-based K-Means, we directly minimize the Euclidean distance between samples and all prototypes. In this case, \SEL{} is shaped as:
\begin{equation}
\mathcal{L}_\mathrm{SEL} = \frac{1}{B}\sum_{i=1}^{B}\sum_{c=1}^{\widetilde{C}} \frac{\exp (f_{\theta}(\bm{x}_i) \cdot \bm{p}_c/\tau)}{\sum\limits_{c=1}^{\widetilde{C}} \exp(f_{\theta}(\bm{x}_i) \cdot \bm{p}_c/\tau)} d(f_\theta(\bm{x}_i), \bm{p}_c),
\label{eq_swl}
\end{equation}
where $\widetilde{C}$ denotes the number of prototypes after merging, e.g., $\widetilde{C}^\mathrm{A}$ is the number of elements in $\mathcal{P}^{\mathrm{A}^\prime}$. By taking all potential semantic correlations into account, \SEL{} can alleviate the impact of the noise within the K-Means clustering results and further guide the model to learn more distinguishable semantic information in terms of Euclidean distance. Certainly, such \SEL{} benefits also rely on high-quality semantic prototypical structures. As a result, we also apply the progressive coefficient $\alpha$ to \SEL{}, then the final optimization objective for \IDSE{} is 
\begin{align}
\mathcal{L}_\mathrm{IDSE} = (\mathcal{L}_\mathrm{IPM}^\mathrm{A} + \mathcal{L}_\mathrm{IPM}^\mathrm{B}) + \alpha (\mathcal{L}_\mathrm{SEL}^\mathrm{A} + \mathcal{L}_\mathrm{SEL}^\mathrm{B}).
\end{align}

\subsection{Cross-Domain Semantic-Matched Learning}
\label{sec_CDSM}
\subsubsection{Semantic-Preserving Domain Alignment.}
\label{sec_spda}
\begin{figure}[htbp]
\centering
\begin{minipage}[h]{0.35\textwidth}
\centering
\includegraphics[width=1.\textwidth]{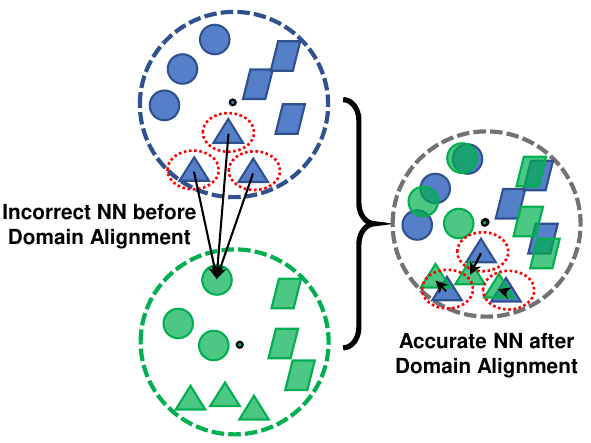}
\vspace{-10pt}
\caption{Comparison of nearest neighbor searching before or after domain alignment.}
\label{fig_incorrect_nn}
\end{minipage}
\quad
\begin{minipage}[h]{0.61\textwidth}
\centering
\includegraphics[width=1.\textwidth]{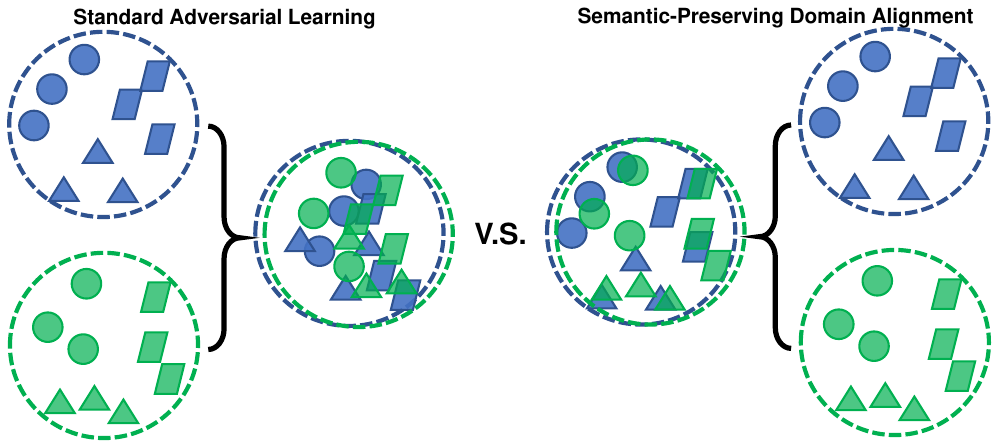}
\vspace{-12pt}
\caption{Comparison between standard adversarial learning and our semantic-preserving domain alignment in terms of semantic structure changes.}
\label{fig_adv_training}
\end{minipage}
\vspace{-15pt}
\end{figure}
Domain invariance is another requirement for the features in \ourproblem{}. However, it is difficult to effectively align feature clusters across domains when no category label nor correspondence annotation can be utilized in our unsupervised setting. There have been some studies that propose a variety of cross-domain categorical instance-matching approaches~\cite{ucdr1, DN2A, PCS} during or after self-supervised pretraining. Specifically, instance matching is proposed to match an instance $\bm{x}_i^\mathrm{A}$ to another instance $\bm{x}_j^\mathrm{B}$ in the other domain with the most similar features. However, due to the domain gap, instances can be easily mapped to instances of different classes in the other domain. In some cases, if there is an instance in one domain that is extremely close to the other domain, it will be determined as the nearest neighbor for all instances in the other domain~\cite{PCS}, shown in Figure~\ref{fig_incorrect_nn}. As a result, before conducting instance matching, the domain gap needs to be diminished. Existing works usually leverage discrepancy minimization~\cite{distributionmatch} or adversarial learning~\cite{DAT} to achieve domain alignment. However, these methods provide inferior performance or have unstable training due to their influence on the semantic categorical structure changes, i.e., the semantic correlations among instances within domains change a great deal during domain alignment, shown in Figure~\ref{fig_adv_training}.

To achieve domain alignment, especially with semantic preservation, we propose Semantic-Preserving Domain Alignment (\SPDA{}). Similar to the standard domain adversarial learning, \SPDA{} happens on two parties, one is the feature extractor $f_\theta$, and the other is a domain classifier $g_\omega$ parameterized on $\omega$. The domain classifier tries to distinguish the representations of two domains, while the feature extractor tries to fool the domain classifier. Thus, the training is shaped like a bi-level optimization in terms of the domain classification task on $\theta$ and $\omega$, shown as follows, 
\begin{align}
\mathcal{L}_\mathrm{DAL} = \sum_{i=1}^{2B} - y_i \cdot \log g_\omega (f_\theta(\bm{x}_i)) - (1-y_i) \cdot \log(1 - g_\omega (f_\theta(\bm{x}_i))),
\end{align}
where $y$ here denotes the domain label, for example, if we regard $y=1$ for domain A, the domain label of domain B is $y=0$. After sufficient adversarial training, the feature extractor captures nearly domain-invariant features, thus achieving domain alignment. 

To prevent the semantic structure from being changed, we first make a copy for the model trained by \IDSE{} and denote it as $f_\theta^\prime$. Then for a mini-batch of a particular domain, we feed all instances to $f_\theta^\prime$ and calculate pair-wise cosine similarity and Euclidean distance. In this way, all these instances constrain and influence each other, which means that when the correlation of a particular instance pair changes, it will affect the correlations of other related pairs. Therefore, we apply a semantic-preserving regulation into domain adversarial learning to make the pair-wise correlations unchanged,
\begin{equation}
\begin{aligned}
\mathcal{L}_\mathrm{SPR} = \frac{1}{B^2} \sum_{i=1}^{B} \sum_{j=1}^{B} &\{\left[\frac{f_\theta(\bm{x}_i) \cdot f_\theta(\bm{x}_j)}{|f_\theta(\bm{x}_i)| |f_\theta(\bm{x}_j)|} - \frac{f_{\theta^\prime}(\bm{x}_i) \cdot f_{\theta^\prime}(\bm{x}_j)}{|f_{\theta^\prime}(\bm{x}_i)| |f_{\theta^\prime}(\bm{x}_j)|}\right]^2 \\
&+ [d(f_\theta(\bm{x}_i), f_\theta(\bm{x}_j)) - d(f_{\theta^\prime}(\bm{x}_i), f_{\theta^\prime}(\bm{x}_j))]^2\}.
\end{aligned}
\end{equation}

\smallskip \noindent \textbf{Recall Semantic-Enhanced Loss.} Actually, aligning two domains together without any semantic structure change is impossible. As a result, there is a need for a dedicated design to alleviate the impact of such unavoidable changes. Our solution is to strengthen the instances' semantic correlations by enhancing the cluster's inner density and inter-separability. As aforementioned, we design \SEL{} that has been incorporated into \IDSE{} learning. For the final convergence of \SEL{}, each instance is optimized to get as close to its corresponding cluster prototype as possible, and as far to other cluster prototypes as possible.

\subsubsection{Switchable Nearest Neighboring Match.}
\label{sec_SNNM}
With \SPDA{}, the domain gap could be effectively minimized and enable more accurate cross-domain instance matching. However, we are afraid that existing instance matching approaches~\cite{ucdr1, PCS, DN2A} cannot take full use of the solid base established by \IDSE{} and \SPDA{}. It is because they lack the capabilities of measuring the matching reliability of an instance and its nearest neighbor with the most similar features. Reasonable matching reliability allows us to conduct cross-domain matching with different weights. For instance, if an instance is located at the joint boundary of multiple categories, which indicates that the current feature extractor is not able to extract sufficient semantic distinguishable features for this instance, thus the reliability of this instance and its nearest neighbor in the other domain is low. In this case, we are supposed to lay less emphasis on unreliable matching instance pairs. To further boost more accurate cross-domain instance matching, we propose the Switchable Nearest Neighboring Match (\SNNM{}).


The principle behind \SNNM{} is that prototypes are more convincing and reliable. Specifically, for a particular sample $\bm{x}_i^\mathrm{A}$ in domain A, we first determine its inner nearest cluster prototype $\bm{p}_{c_i}^\mathrm{A}$ in domain A. Then we can search for the nearest instance $\bm{x}_i^{\mathrm{A}, \mathrm{B}}$ in domain B. Both these two searching processes are based on the product of a modified cosine similarity and Euclidean distance,
\begin{align}
\bm{p}_{c_i}^\mathrm{A} &= \arg\min\limits_{\bm{p}_j^\mathrm{A}} \left[\left(1 - \frac{f_\theta(\bm{x}_i^\mathrm{A}) \cdot \bm{p}_j^\mathrm{A}}{|f_\theta(\bm{x}_i^\mathrm{A}| |\bm{p}_j^\mathrm{A}|}\right) \cdot d(f_\theta(\bm{x}_i^\mathrm{A}), \bm{p}_j^\mathrm{A})\right]\\
\bm{x}_i^{\mathrm{A}, \mathrm{B}} &= \arg\min\limits_{\bm{x}_j^\mathrm{B}} \left[\left(1 - \frac{f_\theta(\bm{x}_i^\mathrm{A}) \cdot f_\theta(\bm{x}_j^\mathrm{B})}{|f_\theta(\bm{x}_i^\mathrm{A}| |f_\theta(\bm{x}_j^\mathrm{B}|}\right) \cdot d(f_\theta(\bm{x}_i^\mathrm{A}), f_\theta(\bm{x}_j^\mathrm{B}))\right].
\end{align}
After obtaining $\bm{x}_i^{\mathrm{A}, \mathrm{B}}$, \SNNM{} searches for its inner nearest prototype $\bm{p}_{c_i}^{\mathrm{A}, \mathrm{B}^\prime}$ in $\mathcal{P}^{\mathrm{B}^\prime}$ (which has been merged with the translated prototype set $\mathcal{P}^{\mathrm{A}\rightarrow\mathrm{B}}$ from domain A) and two cases allows us to measure the reliability of $\bm{x}_i^{\mathrm{A}, \mathrm{B}}$. Before introducing these two cases, the inner nearest prototype $\bm{p}_{c_i}^\mathrm{A}$ of $\bm{x}_i^\mathrm{A}$ needs to be translated to domain B and checked whether should be merged to follow condition Eq.~(\ref{eq_merge_con}), and we denote the translated prototype as $\widetilde{\bm{p}}_{c_i}^\mathrm{A}$. Then the first potential case is $\bm{p}_{c_i}^{\mathrm{A}, \mathrm{B}^\prime}$ is exactly identical to $\widetilde{\bm{p}}_{c_i}^\mathrm{A}$, which means, $\bm{x}_i^{\mathrm{A}, \mathrm{B}}$ is convincing since it shares the same prototype correlations with $\bm{x}_i^\mathrm{A}$ across two domains. Then the pair of $\bm{x}_i^{\mathrm{A}, \mathrm{B}}$ and $\bm{x}_i^\mathrm{A}$ should be viewed as a positive pair in the contrastive loss. Otherwise, if $\bm{p}_{c_i}^{\mathrm{A}, \mathrm{B}^\prime}$ is different from $\widetilde{\bm{p}}_{c_i}^\mathrm{A}$, it may be located at the intersection region of multiple clusters. In this case, the pair of $\bm{x}_i^{\mathrm{A}, \mathrm{B}}$ and $\bm{x}_i^\mathrm{A}$ is not supposed to be treated as a positive pair. However, it does not mean \SNNM{} does nothing for these unreliable cases, instead, \SNNM{} leverages a modified prototype contrastive loss to match the pair of $\bm{x}_i^\mathrm{A}$ and $\widetilde{\bm{p}}_{c_i}^\mathrm{A}$. The modification is to incorporate cross-domain instance-wise negative comparison that can enhance the model generalization ability across domains,
\begin{equation}
\begin{aligned}
&\mathcal{L}_{\mathrm{SN}^2\mathrm{M}} = \frac{1}{B}\sum_{i=1}^{B} -\log \frac{\Delta} {\sum_{c=1}^{\widetilde{C}^\mathrm{B}}\exp(f_\theta(\bm{x}_i^\mathrm{A}) \cdot \bm{p}_{c}^{\mathrm{B}^\prime} / \tau) + \sum_{j=1}^{N^\mathrm{B}}\exp(f_\theta(\bm{x}_i^\mathrm{A}) \cdot f_\theta(\bm{x}_j^\mathrm{B}) / \tau)} \\
&\text{where}\, \Delta = \left\{ 
\begin{array}{ll}
\exp(f_\theta(\bm{x}_i^\mathrm{A}) \cdot \widetilde{\bm{p}}_{c_i}^\mathrm{A} / \tau),\, \text{if}\, \bm{p}_{c_i}^{\mathrm{A}, \mathrm{B}^\prime} \neq \widetilde{\bm{p}}_{c_i}^\mathrm{A}\\
\exp(f_\theta(\bm{x}_i^\mathrm{A}) \cdot \widetilde{\bm{p}}_{c_i}^\mathrm{A} / \tau) + \exp(f_\theta(\bm{x}_i^\mathrm{A}) \cdot f_\theta(\bm{x}_i^{\mathrm{A}, \mathrm{B}}) / \tau), \, \text{otherwise}.
\end{array}
\right.
\end{aligned}
\end{equation}
Finally, the overall optimization objective of \CDSM{} follows
\begin{align}
    \mathcal{L}_\mathrm{CDSM} = \mathcal{L}_\mathrm{DAL} + \mathcal{L}_\mathrm{SPR}^\mathrm{A} + \mathcal{L}_\mathrm{SPR}^\mathrm{B} + \mathcal{L}_{\mathrm{SN}^2\mathrm{M}}^\mathrm{A} + \mathcal{L}_{\mathrm{SN}^2\mathrm{M}}^\mathrm{B}.
\end{align}


%% file: Sections/Experiments.tex
\section{Experiments}
The used datasets, experimental settings, and comparison baselines are introduced below. More implementation details, additional experiment results, and source codes will be provided soon.

\smallskip \noindent \textbf{Datasets.} \emph{Office-31}~\cite{office-31} is composed of three domains with 31 classes: Amazon (A), DSLR (D), Webcam (W). \emph{Office-Home}~\cite{office-home} contains four different domains: Art (A), Clipart (C), Product (P), Real (R). And each domain has 67 data categories. \emph{DomainNet}~\cite{domainnet} is the most challenging cross-domain dataset to our best knowledge, which includes six domains: Quickdraw (Qu), Clipart (Cl), Painting (Pa), Infograph (In), Sketch (Sk) and Real (Re). DomainNet is originally class-imbalanced, thus we follow~\cite{ucdr1} to select data classes that contain more than 200 samples in our following experiments. With this policy, 7 classes are used.

\smallskip \noindent \textbf{Experiment Settings.} For fair comparison, we apply ResNet-50~\cite{resnet} as the backbone of the retrieval model. Following~\cite{ucdr1}, the ResNet-50 model is pre-trained with ImageNet in MoCov2~\cite{moco}. The domain classifier consists of two fully connected layers. The SGD optimizer with a momentum of 0.9 is adopted with an initial learning rate of 0.0002 that is scheduled to zero by a cosine learning strategy. And the batch size is 64. We apply the K-Means and the Elbow approach to construct prototypes and estimate the cluster number, respectively. Following \cite{coda}, we adopt mean average precision on all retrieved results (mAP@All) to measure the performance. All experiments are run repeatedly 3 times with seeds 2024, 2025, and 2026, and we report the mean performance.  

\smallskip \noindent \textbf{Comparison Baselines.} Our proposed method is compared with a comprehensive set of state-of-the-art works from Cross-Domain Representation Learning (CDS~\cite{CDS2021}, PCS~\cite{PCS}), Unsupervised Domain Generalization (DARLING~\cite{darling}, DN$^2$A~\cite{DN2A}), and Unsupervised Cross-Domain Retrieval (UCDIR~\cite{ucdr1}, CoDA~\cite{coda}, DG-UCDIR~\cite{ucdr2}). To test these baselines in different cases of UCDR, we follow their default settings and only conduct compulsory customization.
\begin{table}[h]
\centering
\small
\vspace{-10pt}
\caption{Performance comparison (mAP@All) between ours and other baseline methods on Office-31 and DomainNet in Closet Unsupervised Cross-Domain Retrieval. We blue and bold \blue{\textbf{the best performance}}, and bold \textbf{the second best}, same for all tables.}
\vspace{-10pt}
\resizebox{1\textwidth}{!}{
\setlength{\tabcolsep}{.3mm}{
\begin{tabular}{l|ccccccc|cccccc}
\toprule
\multicolumn{1}{c|}{Methods} & A$\rightarrow$D & A$\rightarrow$W & D$\rightarrow$A & D$\rightarrow$W & W$\rightarrow$A & W$\rightarrow$D & Avg. & Qu$\rightarrow$Cl & Cl$\rightarrow$Pa & Pa$\rightarrow$In & In$\rightarrow$Sk & Sk$\rightarrow$Re & Avg. \\ \midrule
CDS &66.7  &62.5  &70.9  &90.0  &64.4  &88.4  &73.8  &19.2  &35.1  &24.4  &25.5  &32.3  &27.3  \\
PCS &72.7  &70.7  &\textbf{75.3}  &88.5  &71.2  &89.2  &77.9  &22.2  &36.0  &27.7  &\textbf{28.0}  &33.0  &29.4  \\
DARLING &65.5  &70.2  &73.4  &86.6  &69.0  &83.7  &74.7  &22.1  &33.7  &25.9  &27.2  &32.5  &28.3  \\
DN2A &71.1  &\textbf{72.4}  &72.5  &85.8  &71.2  &90.0  &77.2  &23.3  &35.0  &26.0  &27.7  &33.0  &29.0  \\
UCDIR &\textbf{73.7}  &69.9  &74.6  &91.4  &73.1  &90.2  &78.8  &25.8  &\textbf{36.6}  &28.0  &27.5  &\textbf{34.1}  &30.4  \\
CoDA &71.7  &71.4  &74.9  &91.4  &73.1  &90.2  &78.8  &26.0  &34.9  &29.2  &27.9  &33.8  &30.4  \\
DG-UCDIR &73.5  &71.2  &75.1  &\textbf{91.7}  &\textbf{74.0}  &\textbf{90.5}  &\textbf{79.3}  &\textbf{27.7}  &35.0  &\textbf{30.0}  &27.8  &33.6  &\textbf{30.8}  \\ \midrule
Ours &\blue{\textbf{76.2}}  &\blue{\textbf{77.0}}  &\blue{\textbf{75.6}}  &\blue{\textbf{92.5}}  &\blue{\textbf{78.9}}  &\blue{\textbf{91.0}}  &\blue{\textbf{81.9}}  &\blue{\textbf{31.9}}  &\blue{\textbf{39.4}}  &\blue{\textbf{35.0}}  &\blue{\textbf{29.8}}  &\blue{\textbf{35.7}}  &\blue{\textbf{34.4}}  \\ \bottomrule
\end{tabular}}}
\label{tab_closet_office_31}
\end{table}
\begin{table}[h]
\centering
\small
\vspace{-10pt}
\caption{Performance comparison (mAP@All) between ours and other baseline methods on Office-Home in Closet Unsupervised Cross-Domain Retrieval.}
\vspace{-10pt}
\resizebox{1\textwidth}{!}{
\setlength{\tabcolsep}{.3mm}{
\begin{tabular}{l|ccccccccccccc}
\toprule
\multicolumn{1}{c|}{Methods} & A$\rightarrow$C & A$\rightarrow$P & A$\rightarrow$R & C$\rightarrow$A & C$\rightarrow$P & C$\rightarrow$R & P$\rightarrow$A & P$\rightarrow$C & P$\rightarrow$R & R$\rightarrow$A & R$\rightarrow$C & R$\rightarrow$P & Avg. \\ \midrule
CDS &33.0  &44.5  &51.4  &32.4  &40.3  &41.8  &45.3  &41.5  &60.8  &51.1  &42.0  &58.8  &45.2  \\
PCS &34.3  &46.3  &51.6  &32.3  &40.5  &40.6  &47.0  &42.1  &61.3  &51.6  &42.8  &60.1  &45.9  \\
DARLING &32.4  &40.9  &50.5  &33.0  &36.7  &41.5  &47.0  &40.9  &59.0  &51.1  &43.0  &60.8  &44.7  \\
DN2A &35.5  &42.8  &52.9  &34.0  &35.7  &42.0  &48.0  &43.2  &59.8  &49.0  &44.7  &56.5  &45.3  \\
UCDIR &\textbf{36.1}  &46.5  &\textbf{55.9}  &34.0  &44.1  &43.1  &51.2  &44.1  &\textbf{67.1}  &52.7  &43.0  &66.5  &48.7  \\
CoDA &34.7  &49.6  &53.2  &33.2  &42.9  &\textbf{44.7}  &50.4  &45.2  &65.2  &\textbf{53.1}  &\textbf{46.0}  &65.2  &46.8  \\
DG-UCDIR &36.0  &\textbf{50.1}  &55.5  &\textbf{34.1}  &\textbf{44.9}  &43.0  &\textbf{51.5}  &\textbf{45.5}  &66.6  &53.0  &44.5  &\textbf{67.0}  &\textbf{49.3}  \\ \midrule
Ours &\blue{\textbf{38.5}}  &\blue{\textbf{52.6}}  &\blue{\textbf{59.0}}  &\blue{\textbf{34.2}}  &\blue{\textbf{47.5}}  &\blue{\textbf{49.0}}  &\blue{\textbf{55.2}}  &\blue{\textbf{49.0}}  &\blue{\textbf{69.5}}  &\blue{\textbf{56.9}}  &\blue{\textbf{48.7}}  &\blue{\textbf{68.2}}  &\blue{\textbf{52.4}}  \\ \bottomrule
\end{tabular}}}
\label{tab_closet_office_home}
\end{table}
\begin{figure}[h]
\centering
\includegraphics[width=.95\textwidth]{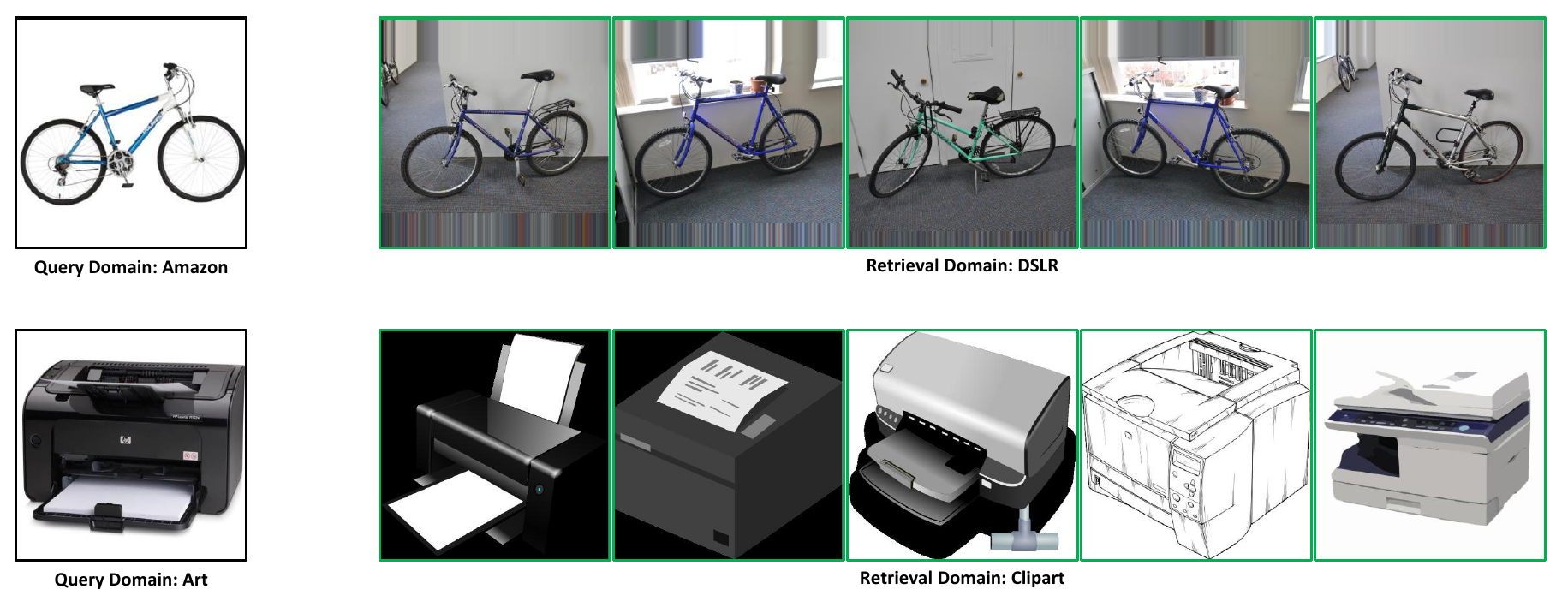}
\vspace{-12pt}
\caption{Retrieval results of our framework on Office-31 (A$\rightarrow$D) and Office-Home (A$\rightarrow$C) in Closet Unsupervised Cross-Domain Retrieval.}
\label{fig:retrieval_res}
\vspace{-5pt}
\end{figure}

\subsection{Effectiveness of \method{} When Solving \ourproblem{}}
\noindent \textbf{Closet Unsupervised Cross-Domain Retrieval.} For the closet setting, the label space of the query domain is the same as that of the retrieval domain. All possible domain pairs of Office-31 and Office-Home are tested, as well as 5 pairs of DomainNet. The detailed experiment results for Office-31, Office-Home, and DomainNet are shown in Tables~\ref{tab_closet_office_31} and \ref{tab_closet_office_home}. According to these results, we can observe that \textit{\textbf{our framework significantly outperforms other baselines in all cases}}. Specifically, we can achieve mAP@All improvement of 2.6\% compared to the second-best baseline method on Office-31, and such improvement is even larger on Office-Home with the range of 3.1\%. On DomainNet, the performance rise can be 3.6\%. Figure~\ref{fig:retrieval_res} also shows the retrieval results of our approach on Office-31 and Office-Home, and we can observe that all results are correct. These results demonstrate that our method can effectively achieve domain-invariant learning with much more accurate cross-domain categorical matching.

\smallskip \noindent \textbf{Partial Unsupervised Cross-Domain Retrieval.} To establish the partial setting, the query domain contains only half of the label space of the retrieval domain, and the query label space is randomly selected. As shown in Tables~\ref{tab_partial_office_31} and \ref{tab_partial_office_home}, we can also observe that \textit{\textbf{\method{} exceeds all other baseline methods a lot in mAP@All}}. Such performance improvement is even larger than that in closet UCDR. For example, \method{} outperforms the second-best with a margin of 12.0\% on Office-31, and the improvement on Office-Home is 12.4\%. Similar improvements on DomainNet can also be observed. In this case, we can easily conclude that existing state-of-the-art studies are nearly incapable of dealing with the label space difference in partial UCDR, while our \method{} can work effectively.
\begin{table}[h]
\centering
\small
\caption{Performance comparison (mAP@All) between ours and other baseline methods on Office-31 and DomainNet in Partial Unsupervised Cross-Domain Retrieval.}
\vspace{-10pt}
\resizebox{1\textwidth}{!}{
\setlength{\tabcolsep}{.3mm}{
\begin{tabular}{l|ccccccc|cccccc}
\toprule
\multicolumn{1}{c|}{Methods} & A$\rightarrow$D & A$\rightarrow$W & D$\rightarrow$A & D$\rightarrow$W & W$\rightarrow$A & W$\rightarrow$D & Avg. & Qu$\rightarrow$Cl & Cl$\rightarrow$Pa & Pa$\rightarrow$In & In$\rightarrow$Sk & Sk$\rightarrow$Re & Avg. \\ \midrule
CDS &42.5  &39.7  &45.4  &\textbf{68.8}  &40.7  &55.8  &48.8  &17.7  &31.1  &22.0  &21.8  &29.6  &24.4  \\
PCS &44.0  &41.5  &47.9  &65.9  &47.3  &56.5  &50.5  &18.0  &\textbf{31.5}  &23.4  &22.1  &27.9  &24.6  \\
DARLING &41.5  &38.9  &49.0  &65.5  &45.7  &53.6  &49.0  &19.0  &30.5  &23.3  &23.0  &28.7  &24.9  \\
DN2A &42.5  &40.0  &\textbf{51.1}  &66.7  &46.5  &55.0  &50.3  &18.7  &31.1  &25.0  &24.4  &30.7  &26.0  \\
UCDIR &\textbf{46.0}  &\textbf{41.8}  &50.3  &62.9  &47.0  &56.9  &50.8  &19.3  &29.8  &24.4  &23.1  &29.0  &25.1  \\
CoDA &45.1  &40.7  &49.3  &66.0  &47.0  &\textbf{57.8}  &\textbf{51.0}  &\textbf{20.2}  &30.9  &\textbf{25.2}  &24.0  &31.0  &\textbf{26.3}  \\
DG-UCDIR &45.0  &39.2  &49.7  &64.4  &\textbf{48.5}  &55.2  &50.3  &18.8  &30.4  &25.0  &\textbf{24.5}  &\textbf{31.2}  &26.0  \\ \midrule
Ours &\blue{\textbf{64.4}}  &\blue{\textbf{51.0}}  &\blue{\textbf{59.4}}  &\blue{\textbf{76.9}}  &\blue{\textbf{61.5}}  &\blue{\textbf{65.0}}  &\blue{\textbf{63.0}}  &\blue{\textbf{28.2}}  &\blue{\textbf{34.9}}  &\blue{\textbf{32.7}}  &\blue{\textbf{26.6}}  &\blue{\textbf{34.0}}  &\blue{\textbf{31.3}}  \\ \bottomrule
\end{tabular}}}
\label{tab_partial_office_31}
\vspace{-10pt}
\end{table}
\begin{table}[h]
\centering
\small
\caption{Performance comparison (mAP@All) between ours and other baseline methods on Office-Home in Partial Unsupervised Cross-Domain Retrieval.}
\vspace{-10pt}
\resizebox{1\textwidth}{!}{
\setlength{\tabcolsep}{.3mm}{
\begin{tabular}{l|ccccccccccccc}
\toprule
\multicolumn{1}{c|}{Methods} & A$\rightarrow$C & A$\rightarrow$P & A$\rightarrow$R & C$\rightarrow$A & C$\rightarrow$P & C$\rightarrow$R & P$\rightarrow$A & P$\rightarrow$C & P$\rightarrow$R & R$\rightarrow$A & R$\rightarrow$C & R$\rightarrow$P & Avg. \\ \midrule
CDS &22.0  &31.1  &32.5  &26.5  &25.6  &27.9  &30.0  &31.8  &40.5  &32.3  &25.5  &37.6  &30.3  \\
PCS &24.5  &36.5  &38.8  &24.9  &28.8  &29.0  &28.6  &\textbf{35.3}  &41.7  &\textbf{37.5}  &26.9  &40.0  &32.7  \\
DARLING &25.5  &34.7  &29.8  &25.0  &23.9  &27.5  &26.8  &31.9  &40.0  &35.5  &27.7  &40.0  &30.7  \\
DN2A &\textbf{25.9}  &\textbf{37.0}  &29.5  &25.2  &27.0  &\textbf{30.5}  &29.0  &31.5  &40.6  &35.7  &28.0  &41.0  &31.7  \\
UCDIR &23.0  &28.7  &31.0  &26.0  &22.0  &23.5  &\textbf{31.1}  &30.4  &40.2  &36.9  &27.0  &36.8  &29.7  \\
CoDA &22.5  &34.2  &35.7  &25.0  &29.5  &30.0  &30.7  &32.0  &43.2  &35.2  &28.5  &41.3  &32.3  \\
DG-UCDIR &24.4  &30.9  &\textbf{41.0}  &\textbf{27.2}  &\textbf{30.5}  &29.6  &30.4  &33.2  &\textbf{45.5}  &37.1  &\textbf{30.9}  &\textbf{42.0}  &\textbf{33.6}  \\ \midrule
Ours &\blue{\textbf{40.5}}  &\blue{\textbf{45.8}}  &\blue{\textbf{48.0}}  &\blue{\textbf{35.1}}  &\blue{\textbf{39.2}}  &\blue{\textbf{41.1}}  &\blue{\textbf{52.4}}  &\blue{\textbf{46.0}}  &\blue{\textbf{55.0}}  &\blue{\textbf{49.0}}  &\blue{\textbf{43.1}}  &\blue{\textbf{56.7}}  &\blue{\textbf{46.0}}  \\ \bottomrule
\end{tabular}}}
\label{tab_partial_office_home}
\end{table}

\smallskip \noindent \textbf{Open-set Unsupervised Cross-Domain Retrieval.} As for the open-set setups, we ensure the label space of the retrieval domain is half of the query label space. The experiment results for DomainNet are presented in Table~\ref{tab_open_domainnet} with two metrics -- the mAP@All for the shared label set, and the detection accuracy for the private (open-set) query labels (please refer to materials we will provide in the future for details of how to determine whether a query sample belongs to private query labels). According to these results, we can easily observe that \textit{\textbf{our approach substantially exceeds other baseline methods in both two metrics}}. Similar trends can also be found for Office-31 and Office-Home. All these results strongly validate the effectiveness of \method{} in open-set UCDR. 
\begin{table}[ht]
\centering
\begin{minipage}{0.55\linewidth}
\centering
\small
\caption{Performance comparison (mAP@All for shared-label set, detection accuracy for open-label set) between ours and other baseline methods on DomainNet in Openset Unsupervised Cross-Domain Retrieval.}
\vspace{-8pt}
\resizebox{1.\textwidth}{!}{
\setlength{\tabcolsep}{.6mm}{
\begin{tabular}{l|cc|cc|cc|cc|cc|cc}
\toprule
\multicolumn{1}{c|}{\multirow{2}{*}{Methods}} & \multicolumn{2}{c}{Qu$\rightarrow$Cl} & \multicolumn{2}{c}{Cl$\rightarrow$Pa} & \multicolumn{2}{c}{Pa$\rightarrow$In} & \multicolumn{2}{c}{In$\rightarrow$Sk} & \multicolumn{2}{c}{Sk$\rightarrow$Re} & \multicolumn{2}{c}{Avg.} \\ \cmidrule{2-13} 
\multicolumn{1}{c|}{} & \multicolumn{12}{c}{Shared-set mAP@All / Open-set Acc} \\ \midrule
CDS &22.4  &58.9  &34.5  &65.2  &25.5  &60.7  &25.0  &59.2  &33.7  &64.9  &28.2  &61.8  \\
PCS &23.3  &57.8  &34.2  &67.8  &24.9  &60.5  &\textbf{27.8}  &\textbf{65.4}  &34.7  &66.9  &29.0  &63.7  \\
DARLING &21.9  &54.4  &32.5  &60.2  &22.0  &53.9  &26.6  &60.6  &32.3  &62.8  &27.1  &58.4  \\
DN2A &22.7  &56.6  &33.4  &60.7  &21.9  &55.2  &24.8  &57.8  &34.0  &61.2  &27.4  &58.3  \\
UCDIR &\textbf{24.4}  &\textbf{59.0}  &34.4  &67.0  &26.7  &\textbf{62.9}  &25.6  &63.4  &\textbf{35.5}  &\textbf{68.2}  &\textbf{29.3}  &\textbf{64.1}  \\
CoDA &24.2  &58.8  &35.6  &66.6  &\textbf{27.0}  &61.1  &24.9  &58.0  &34.6  &62.9  &\textbf{29.3}  &61.5  \\
DG-UCDIR &23.5  &57.5  &\textbf{36.0}  &\textbf{68.3}  &25.8  &60.6  &25.8  &59.1  &35.0  &64.3  &29.2  &62.0  \\ \midrule
Ours &\blue{\textbf{30.3}}  &\blue{\textbf{72.9}}  &\blue{\textbf{40.2}}  &\blue{\textbf{88.1}}  &\blue{\textbf{36.0}}  &\blue{\textbf{82.5}}  &\blue{\textbf{31.1}}  &\blue{\textbf{78.2}}  &\blue{\textbf{36.6}}  &\blue{\textbf{83.0}}  &\blue{\textbf{34.8}}  &\blue{\textbf{80.9}}  \\ \bottomrule
\end{tabular}}}
\label{tab_open_domainnet}
\end{minipage}
\quad
\begin{minipage}{0.41\linewidth}
\centering
\small
\caption{Ablation studies of UEM on Office-31, Office-Home, and DomainNet in Openset Unsupervised Cross-Domain Retrieval. The average values of Shared-set mAP@All and Open-set Acc for all domain pairs are reported here.}
\vspace{-10pt}
\resizebox{1\textwidth}{!}{
\setlength{\tabcolsep}{.5mm}{
\begin{tabular}{l|cccccc}
\toprule
\multicolumn{1}{c|}{\multirow{2}{*}{Variations}} & \multicolumn{2}{c|}{$\,\,\,\,\,$Office-31$\,\,\,\,\,$} & \multicolumn{2}{c|}{Office-Home} & \multicolumn{2}{c}{DomainNet} \\ \cmidrule{2-7} 
 & \multicolumn{6}{c}{Shared-set mAP@All / Open-set Acc} \\ \midrule
Ours w/o P.M. &68.3  & \multicolumn{1}{c|}{78.8} &45.7  & \multicolumn{1}{c|}{72.6} &30.9  &70.2  \\
Ours w/o \SEL{} &72.2  & \multicolumn{1}{c|}{88.3} &47.5  & \multicolumn{1}{c|}{81.7} &33.0  &76.4  \\
Ours w/o \SPDA{} &62.2  & \multicolumn{1}{c|}{61.9} &39.0  & \multicolumn{1}{c|}{60.8} &24.4  &60.5  \\
Ours w/ UCDIR &75.0  & \multicolumn{1}{c|}{89.2} &48.8  & \multicolumn{1}{c|}{82.8} &33.3  &80.5  \\
Ours &77.4  & \multicolumn{1}{c|}{92.5} &50.2  & \multicolumn{1}{c|}{86.7} &34.8  &80.9  \\ \bottomrule
\end{tabular}}}
\label{tab_ablation}
\end{minipage}
\vspace{-10pt}
\end{table}

\subsection{Ablation Study}
All the ablation studies are carried out in open-set UCDR on three datasets, and the average metrics for all domain pairs of a single dataset are reported here.

\smallskip \noindent \textbf{Effectiveness of Prototype Merging.} When evaluating the effectiveness of building a unified prototypical structure across domains, we don't use prototype merging in \IDSE{}. According to the results of `Ours w/o P.M.' in Table~\ref{tab_ablation}, we can see that there is a non-negligible performance drop in both shared-set mAP@All and open-set accuracy. This validates the importance of building a unified prototypical structure across domains.

\smallskip \noindent \textbf{Effectiveness of \SEL{}.} When evaluating \SEL{}, we just detach and don't use it during the model training. According to the results of `Ours w/o \SEL{}' in Table~\ref{tab_ablation}, there is also an evident performance drop compared to the full \method{}. This validates that \SEL{} is vital as it can help prepare a better base model for \CDSM{}.

\smallskip \noindent \textbf{Effectiveness of \SPDA{}.} We replace our \SPDA{} with the standard domain adversarial learning for the ablation study. As shown in Table~\ref{tab_ablation}, there is a significant performance difference between `Ours w/o \SPDA{}' and the full \method{}, which illustrates the importance of semantic preservation during domain alignment, as well as indirectly verifying the necessity of \SPDA{}.

\smallskip \noindent \textbf{Effectiveness of \SNNM{}.} We also replace the proposed \SNNM{} with another semantic similarity-based nearest neighboring search approach, which is leveraged by UCDIR. By comparing the experiment results of `Ours w/ UCDIR' and `Ours', we can conclude that \SNNM{} is more compatible with \method{} and able to achieve more accurate cross-domain categorical matching.

%% file: Sections/Conclusion.tex
\section{Conclusion}
In this work, we focus on two major challenges when employing cross-domain retrieval (CDR) in real-world scenarios: one is that the category space across domains is usually distinct, and the other is both the query and retrieval domains are unlabeled. To tackle these challenges, the proposed Unified, Enhanced, and Matched (\method{}) semantic feature learning framework can establish a unified semantic structure across domains and preserve this structure during categorical matching. Extensive experiments in cases including closet, partial, open-set unsupervised CDR on multiple datasets demonstrate the effectiveness and universality of \method{}, which are reflected in the substantial performance improvement over state-of-the-art studies from Cross-Domain Representation Learning, Unsupervised Domain Generalization, and Unsupervised CDR.